%% file: main.tex
\documentclass[conference]{IEEEtran}
\IEEEoverridecommandlockouts
\usepackage{cite}
\usepackage{amsmath,amssymb,amsfonts}
\usepackage{algorithmic}
\usepackage{graphicx}
\usepackage{textcomp}
\usepackage{xcolor}
\usepackage{authblk}
\usepackage{multirow}
\usepackage{booktabs}
\usepackage{hyperref}
\usepackage{authblk}
\usepackage{subcaption}
\usepackage[export]{adjustbox}
\usepackage[font=small,labelfont=bf]{caption}
\usepackage[ruled,vlined]{algorithm2e}
\hypersetup{
    colorlinks=true,
    linkcolor=blue,
    filecolor=magenta,      
    urlcolor=cyan,
    citecolor = magenta,
    pdfpagemode=FullScreen,
}

\newcommand{\blue}[1]{\textcolor{blue}{#1}}

\makeatletter
\newcommand*\titleheader[1]{\gdef\@titleheader{#1}}
\AtBeginDocument{%
  \let\st@red@title\@title
  \def\@title{%
    \bgroup\normalfont\large\centering\@titleheader\par\egroup
    \vskip1em\st@red@title}
}
\makeatother

\def\BibTeX{{\rm B\kern-.05em{\sc i\kern-.025em b}\kern-.08em
    T\kern-.1667em\lower.7ex\hbox{E}\kern-.125emX}}

\title{
\vspace{-0.05in}
FRL-FI: Transient Fault Analysis for Federated Reinforcement Learning-Based Navigation Systems
}
\titleheader{2022 Design Automation and Test in Europe Conference (DATE)}
\author[]{Zishen~Wan$^1$, Aqeel~Anwar$^1$, Abdulrahman~Mahmoud$^2$, Tianyu~Jia$^{3,4}$, Yu-Shun~Hsiao$^2$, 
\\Vijay~Janapa~Reddi$^2$, Arijit~Raychowdhury$^1$\vspace{0.01in}
\\ $^1$Georgia Institute of Technology, $^2$Harvard University, $^3$Carnegie Mellon University, $^4$Peking University \vspace{-0.05in}}

\begin{document}


\maketitle

\begin{abstract}
Swarm intelligence is being increasingly deployed in autonomous systems, such as drones and unmanned vehicles. Federated reinforcement learning (FRL), a key swarm intelligence paradigm where agents interact with their own environments and cooperatively learn a consensus policy while preserving privacy, has recently shown potential advantages and gained popularity. However, transient faults are increasing in the hardware system with continuous technology node scaling and can pose threats to FRL systems. Meanwhile, conventional redundancy-based protection methods are challenging to deploy on resource-constrained edge applications. In this paper, we experimentally evaluate the fault tolerance of FRL navigation systems at various scales with respect to fault models, fault locations, learning algorithms, layer types, communication intervals, and data types at both training and inference stages. We further propose two cost-effective fault detection and recovery techniques that can achieve up to 3.3$\times$ improvement in resilience with $<$2.7\% overhead in FRL systems.

\end{abstract}

\input{1introduction}
\input{2related}
\input{3methodology}
\input{4results}
\input{5mitigation}
\input{6conclusion}
\bibliographystyle{ieeetr}
\bibliography{refs}

\end{document}

%% file: 1introduction.tex
\section{Introduction}
\label{sec:intro}
Federated Reinforcement Learning (FRL) is increasingly attracting attention and being adopted in autonomous navigation systems~\cite{chakraborty2017swarm,anwar2021multi}. Multiple agents interact with their own environments and collaboratively learn a unified policy by only sharing the policy or gradient information. FRL improves overall performance while preserving the privacy of data being collected locally and reducing the communication cost.

However, practical reliability considerations pose  challenges to the widespread real-life deployment of FRL systems. Adversarial attacks~\cite{tolpegin2020data}, software failures~\cite{lyu2020threats}, and communication failures~\cite{ang2020robust} can severely violate FRL task safety. Particularly, with the continuous technology node scaling and the lowering of power supply voltage, transient faults, such as soft errors, are increasing in the compute systems, and this impact may exacerbate with the booming accelerators for autonomous machine computing~\cite{wan2021survey,liu2021robotic,gao2021ielas,wan2022circuit}. \cite{jha2019ml,hsiao2021mavfi,wan2021analyzing} demonstrate that such errors can significantly impact the reliability of safety-critical autonomous vehicles.
However, the resilience of FRL navigation systems to transient faults is not well studied. Thus, there is a strong need to understand the vulnerability of FRL systems and develop effective protection techniques.

Transient faults may greatly impact FRL in both training and inference. In FRL training, multiple agents interact with each other. Different from conventional offline training and direct deployment methods, FRL requires real-time training and fine-tuning due to the disparity between the simulated and real environments. Faults in training might impact system convergence and derail the learning process. In FRL inference, agents' actions are made by a long-term sequential decision-making process rather than a single non-sequential DNN in supervised learning. Faults in one stage might propagate to the following stages and impact the end-to-end system resilience.



Conventional techniques to protect a system from hardware faults usually consist of  redundant-based hardware~\cite{hudson2018fault} or ECC~\cite{park2013vl}. While these methods are effective, they bring a large overhead in resource-constrained edge devices. Besides, adding these schemes to all agents in FRL may lead to over-design of the overall system. Therefore, we need a suitable fault mitigation technique for edge applications while considering the distributed nature of the system.

In this work, we present FRL-FI, an in-depth fault characterization on FRL systems from small (a template grid-based navigation task) to large (a drone navigation task) computing scales. The resilience of the system is evaluated by injecting transient faults in both training and inference. We explore how learning algorithms, fault locations, agent numbers, data types, and agent-server communication impact the systems' resilience and performance. We demonstrate the higher resilience of FRL systems compared to systems with a single-agent. Based on the observation that faults in the server have a larger impact than that in agents, we propose a cost-effective fault mitigation technique: checkpointing only in the server during training and using range-based anomaly detection during inference.

In summary, this paper makes the following contributions:
\begin{itemize}
    \item Development of an end-to-end reliability analysis framework to analyze the fault tolerance of FRL systems.
    
    \item Exploration of the impact of large-scale transient fault injection in both training and inference of FRL navigation systems, concerning fault location, communication, and agent numbers. To the best of our knowledge, this is the first work on hardware fault analysis in FRL systems.
    
    \item Improvement  of the FRL system reliability, by proposing low-overhead fault mitigation scheme and evaluating it across various application scenarios. By checkpointing in the server and detecting anomaly values in agents, the system's resilience is improved by up to 3.3$\times$.
\end{itemize}

%% file: 2related.tex
\section{Related Work}
\label{sec:related} 
\subsection{Reliability of Federated Reinforcement Learning Systems}
The effects of various attacks or faults have been evaluated in FRL systems. Data-poisoning and policy-poisoning attacks are evaluated by feeding in false data or policy~\cite{anwar2021multi,tolpegin2020data,lyu2020threats}, but the robustness of FRL to hardware faults has not been adequately explored.
\cite{reagen2018ares,schorn2019efficient,mahmoud2020pytorchfi} assess the impact of hardware faults in single neural network. However, the reliability of FRL depends on how faults propagate in sequential decision-making tasks and multiple agents. This paper aims at investigating the reliability of FRL systems under these challenges.

\subsection{Reliability of Navigation Systems}


The reliability of navigation systems has been recently studied in the field of autonomous vehicles. Toschi et al.~\cite{toschi2019characterizing} study the sensor noise impacts, while we evaluate the hardware faults in compute. DriveFI~\cite{jha2019ml} and MAVFI~\cite{hsiao2021mavfi} assess transient errors on conventional model-based autonomy paradigms consisting of several modules such as perception, planning, and localization. Wan et al.~\cite{wan2021analyzing} study the fault tolerance of single-agent learned-based navigation systems. However, the resilience of multi-agent end-to-end learning-based systems
to hardware faults is still not well understood.




\subsection{Fault Mitigation Techniques}
Several techniques have been proposed to mitigate faults, including DMR, TMR~\cite{hudson2018fault}, and ECC~\cite{park2013vl}. However, these techniques incur large power and area overhead and are challenging to deploy on resource-constrained edge nodes (e.g., drones). Checkpoint-based recovery schemes~\cite{pena2015vocl,akturk2020acr} are effective for fault mitigation, but they are usually deployed in high performance computing. In this work, we propose an application-aware symptom-based checkpoint-based fault protection scheme with low end-to-end system overhead, while considering the characteristic of the distributed system and its inherent resilience.








%% file: 3methodology.tex
\vspace{-0.01in}
\section{Methodology and Fault Model} 
\label{sec:meth}
This section introduces the federated reinforcement learning navigation systems (\blue{\S}\ref{subsec:RL_methoodology}, \blue{\S}\ref{subsec:tarining_inference}). 
We present the fault model in \blue{\S}\ref{subsec:fault_model} 
and fault injection methodology in \blue{\S}\ref{subsec:fault_injection_method}.

\subsection{Federated Reinforcement Learning (FRL)}
\label{subsec:RL_methoodology}
\begin{figure}[t]
        \centering\includegraphics[width=.75\columnwidth]{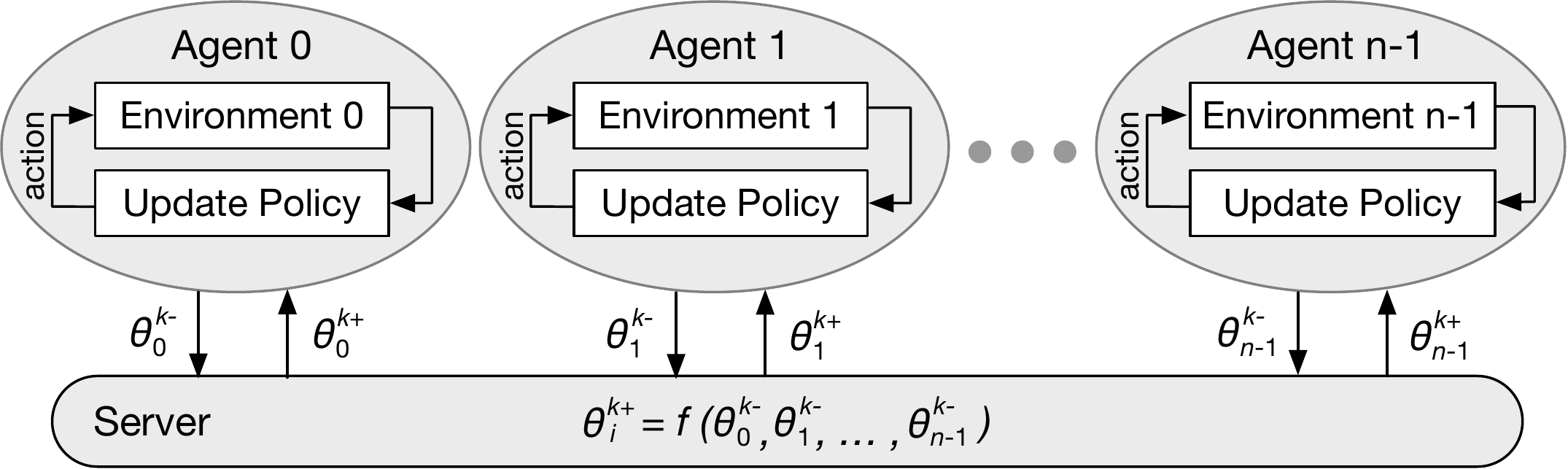}
        \caption{Overview of FRL System. All agents learn a unified policy without sharing the local data that works good for all environments.}
        \vspace{-0.2in}
        \label{fig:FRL}
\end{figure}

In this work, we focus on reliability analysis of FRL-based navigation systems (Fig.~\ref{fig:FRL}).
Consider a FRL problem with $n$ agents operating in $n$ different environments and are responsible for executing their own tasks.
The tuple $\mathcal{M}_i = (\mathcal{S}_i,\mathcal{A}_i,\mathcal{P}_i,\mathcal{R}_i,\gamma_{i})$ can be used to describe the Markov Decision Process (MDP) for each environment $i$, where $\mathcal{S}_i$ is the state space, $\mathcal{A}_{i}$ is the action space, $\mathcal{P}_{i}$ is the MDP transition probabilities, $\mathcal{R}_{i}:\mathcal{S}_{i}\times\mathcal{A}_{i}\rightarrow \mathbb{R}$ is the reward function that directly qualifies the nature of the underlying task in environment $i$, and $\gamma_{i}\in(0,1)$ is the discount factor. 

Let $V^{\pi}_i$ be the value function, at the state $s$ in the $i$-th environment, induced by the policy $\pi$. Then, we have
\begin{align}
&V_i^{\pi}(s) = \mathbb{E}\left[\sum_{k=0}^{\infty}\gamma_{i}^{k}\mathcal{R}_{i}(s^{k}_{i},a_{i}^{k})\,|\,s_{i}^{0} = s\right],\,\,
\label{sec:prob:value_function}  
\end{align}

where $a_{i}^{k} \sim \pi(\cdot|s_{i}^{k})$. We use $\rho_{i}$ to denote the initial state distribution over the action space of the $i$-th environment.
The objective of the FRL problem is to find a unified policy $\pi^*$ that maximizes the sum of long-term discounted return, quantified by the value function $V_i^{\pi}(s)$ for all $i$ environments:
\begin{align}
\max_{\pi} V(\pi;\boldsymbol{\rho}) \triangleq \sum_{i=0}^{n-1}\mathbb{E}_{s_{i}\sim\rho_{i}}V_{i}^{\pi}(s_{i}),
\label{eq:unified_objective}
\end{align}

where $\boldsymbol{\rho}=[\rho_0, \ldots, \rho_{n-1}]^T$. Solving Eq.~\ref{eq:unified_objective} yields a unified $\pi^*$ resulting in a balanced performance across $n$ environments. 

We use $\theta$ to model the family of policies $\pi_\theta(a|s)$, then the goal of the FRL problem is to find $\theta^*$ that satisfies
\begin{align}
\theta^* = \arg\max_{\theta} V(\theta;\boldsymbol{\rho}) \triangleq \sum_{i=0}^{n-1}\mathbb{E}_{s_{i}\sim\rho_{i}}V_{i}^{\pi_{\theta}}(s_{i}).   \label{sec:prob:obj_theta}
\end{align}


To find the unified policy in Eq. \ref{eq:unified_objective}, in FRL, $\mathcal{M}_i$ remains at the local agent $i$ while the policy $\theta_i$ is shared with a common designated agent (\textit{server}). Each agent $i$ tries to learn its own task, by utilizing its local data $\mathcal{D}_i$ to train $\theta_i$.
After completing each episode $k$, agents share their policy $\theta_i^{k-}$ with the server. The server carries out a smoothing average and generates $n$ new sets of parameters $\theta^{k+}_i$, one for each agent, using 
$
    \theta^{k+}_i = \alpha^k \theta_i^{k-} + \beta^k \sum_{j \neq i} \theta_j^{k-}
    \label{eq:smooth_avg} $
where $\alpha^k, \beta^k=\frac{1-\alpha}{n-1} \in (0, 1)$ are smoothing average weights. The goal of this smoothing average is to achieve a consensus among the agents' parameters, i.e.
\begin{equation}
    \lim_{k\rightarrow \infty} \theta^{k+}_i \rightarrow \theta^*  \quad \forall i \in \{0, n-1\}.
    \label{eq:convergence}
\end{equation}

As the training proceeds, the smoothing average constants are guaranteed to converge to $\alpha^k, \beta^k \rightarrow \frac{1}{n}$~\cite{zeng2021decentralized}. 

To summarize, in the FRL system, each agent interacts with its own environments and updates policy parameters, and finally all agents can converge to learn a unified policy.





\subsection{FRL Training and Inference Phases}
\label{subsec:tarining_inference}
In this paper, we analyze both training and inference. 
Policy is first trained on meta environments, and then transferred to real scenarios with fine-tuning because the sim-to-real discrepancy may degrade performance~\cite{anwar2020autonomous}.
The ratio of exploration to exploitation decreases during this model tuning, and agent will finally consistently conduct exploitation when the policy performs within the target error bounds. Two main phases in the on-device procedure are (1) phase with changing exploration-exploitation ratio (\textit{training}) and (2) a greedy exploitation phase (\textit{inference}). Faults in each phase impact the final performance.

\subsection{Fault Model}
\label{subsec:fault_model}
\textbf{Fault type.}
We consider transient faults in this work, which occur from external disturbances and may only exist for a short period. We use a widely employed random bit-flip model~\cite{mittal2020survey} as an abstraction of physical defect mechanisms in devices and systems. Single or multiple bits in data or memory elements are randomly flipped, thus emulating incorrect data capture.



\textbf{Fault location in the FRL system.}
We consider three fault sources: server, communication, and agent. In communication, we consider transient faults due to interference, distortion or synchronization problems leading to incorrect shared parameters. In server and agent, we consider transient faults in memory, including weights, feature maps, and activations. We target edge applications where computation mainly happens in hardware accelerators. We do not consider faults in control logic since scheduling is mainly done by the host CPU.
Our fault model aligns with previous work in this field~\cite{li2017understanding,reagen2018ares}. 

To simplify the analysis (\blue{\S}\ref{sec:experiments}), we group these three fault sources into two classes: faults in the data that agents receive (including faults in server and server-to-agent communication), and faults in the data that the server receives (including faults in agents and agent-to-server communication). We use \textit{agent faults} and \textit{server faults} respectively in following sections.

\textbf{Bit Error Rate (BER).}
We start with single bit-flip and gradually increase BER in experiments. With continuous technology node scaling and lowering of the operating supply in edge devices, \cite{stutz2021bit} shows hardware BER tends to increase up to $>$$10^{-2}$ in 14nm SRAM. \cite{massoud2007digital} shows wireless communication BER can increase up to $>$$10^{-2}$ with decreasing channel SNR. Conventional computing systems typically require BER$<$$10^{-15}$ to guarantee correctness~\cite{BER2017}, levying high design cost. However, we demonstrate that learning-based systems are more robust than conventional workloads. Relaxing BER requirement can enable significant savings and provide opportunities for efficiency improvement across device and architecture levels. 



\subsection{Fault Injection Methodology}
\label{subsec:fault_injection_method}
Faults can be injected and simulated at different levels of the system stack. FRL system usually has long simulation time, and \cite{jha2019ml,hsiao2021mavfi} indicate that software-level injection is the most suitable for such end-to-end autonomous navigation system evaluation due to its fast speed and acceptable accuracy.
We use two software-level fault injection modes: static injection and dynamic injection. Static injection is performed before inference execution begins and introduce zero runtime overhead. Dynamic injection is conducted during training or inference, and the overheads are minimized by implementing fault models as native tensor operations. Faults in the model weights belong to static injection during inference since they are fixed once trained, while activations and weight faults in training belong to the dynamic injection class since they are updated at each step.
This fault injection method aligns with~\cite{reagen2018ares,schorn2019efficient}, where the accuracy of the schemes have been validated on silicon~\cite{whatmough201714}.


%% file: 4results.tex
\vspace{-0.01in}
\section{Fault Characterization and Analysis}

\label{sec:experiments}
We evaluate the fault tolerance of FRL on both a simple, grid-based navigation system (\blue{\S}\ref{subsec:gridworld}) as well as a complex, drone autonomous navigation system (\blue{\S}\ref{subsec:drone}).

\begin{figure}[t]
        \centering\includegraphics[width=\columnwidth]{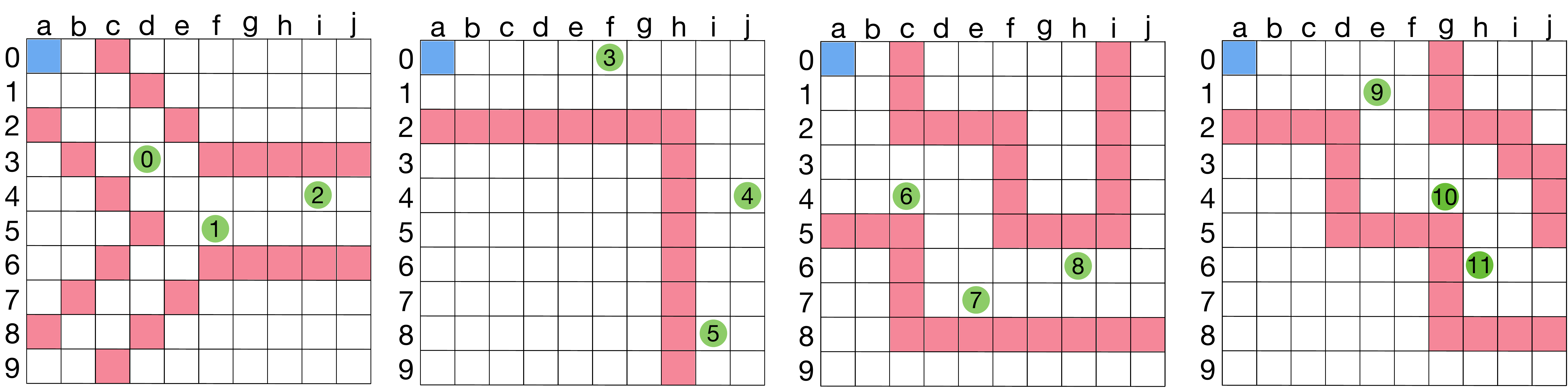}
        \caption{[GridWorld] Grid-based FRL problem, where we combine 12 environemnts into 4 grids. Blue, green and red cells represent agent, goals and obstacles, respectively. For each environment, we terminate the test when agent reaches the goal or hits an obstacle.}
        \vspace{-0.2in}
        \label{fig:gridworld}
\end{figure}

\begin{figure*}[t!]
\centering
    \begin{subfigure}[t]{.24\linewidth}
        \includegraphics[width=\textwidth]{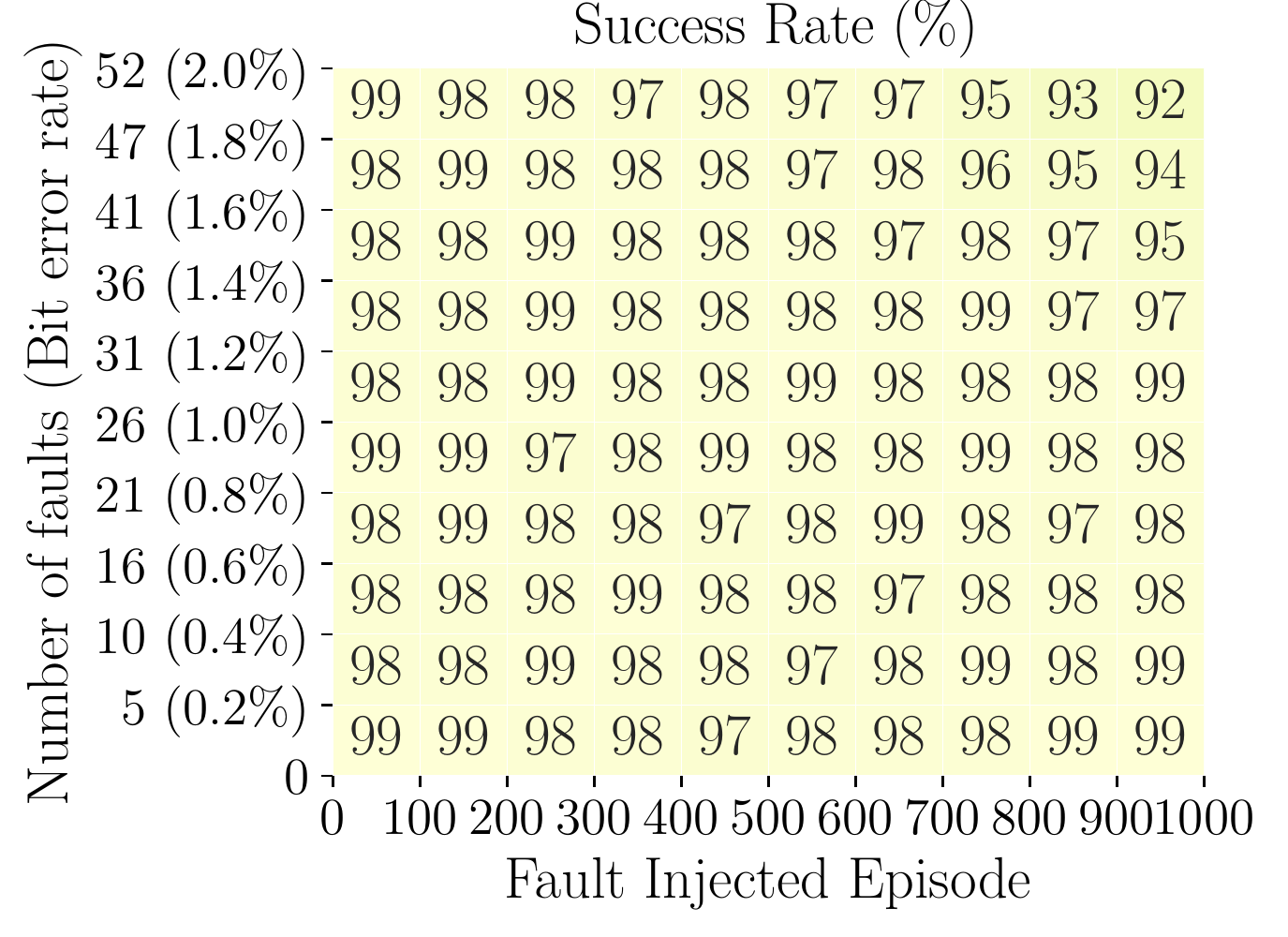}
        \vspace{-0.2in}
        \caption{FRL: agents faults}
        \label{subfig:NN_agent}
    \end{subfigure}
    \hspace{0.005in}
    \begin{subfigure}[t]{.173\linewidth}
        \includegraphics[width=\textwidth]{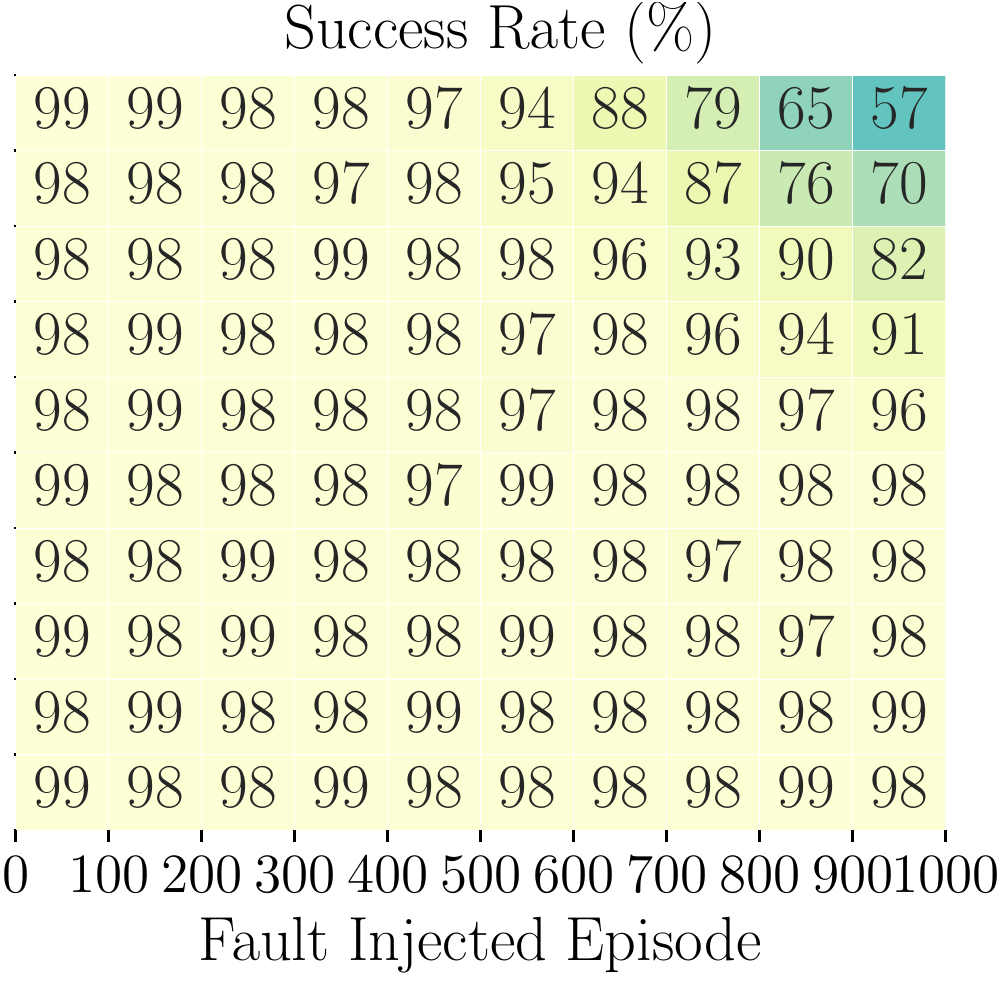}
        \vspace{-0.2in}
        \caption{FRL: server faults}
        \label{subfig:NN_server}
    \end{subfigure}
    \hspace{0.005in}
    \begin{subfigure}[t]{.21\linewidth}
        \includegraphics[width=\textwidth]{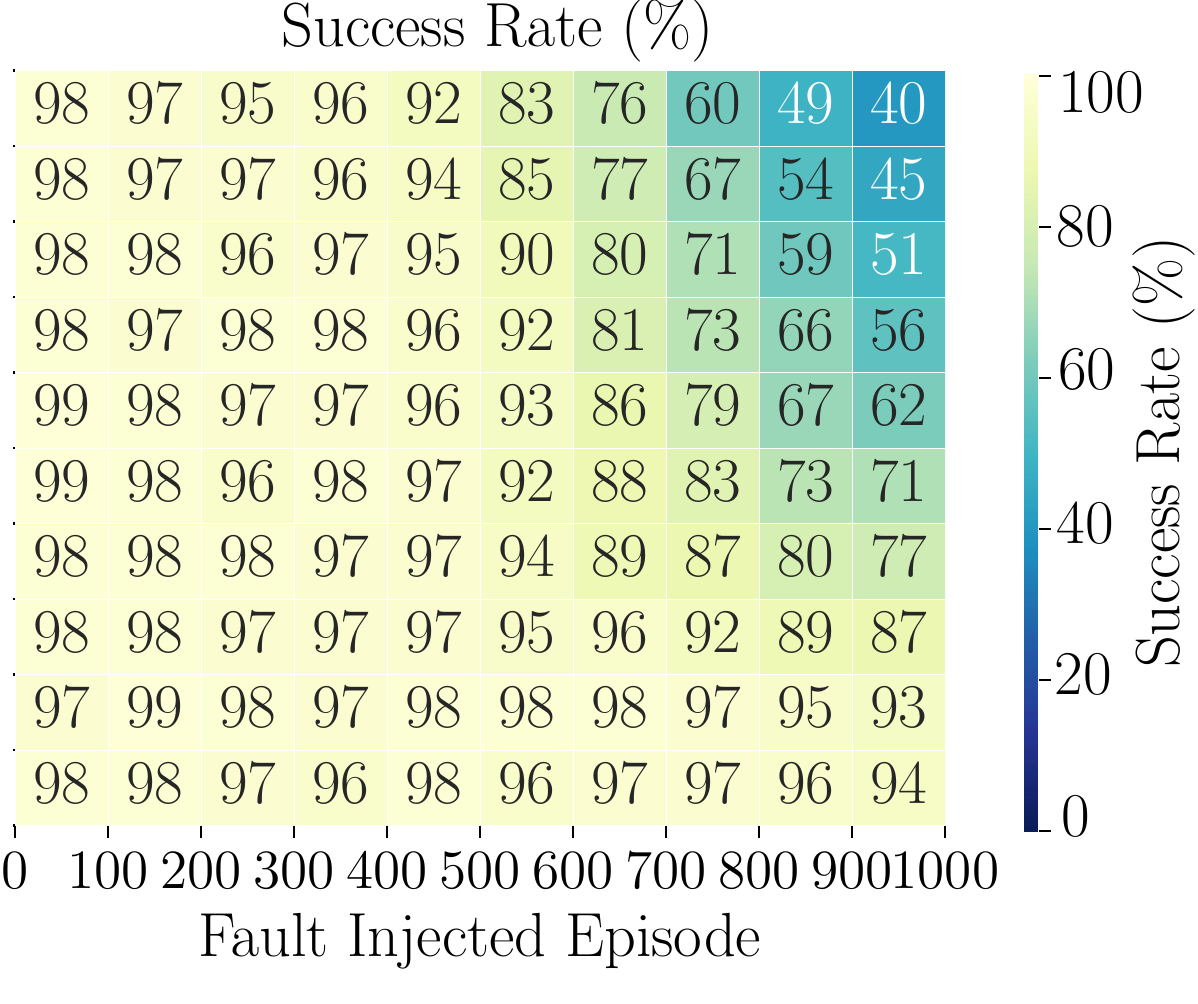}
        \vspace{-0.2in}
        \caption{Single-agent (no server)}
        \label{subfig:single_agent}
    \end{subfigure}
    \hspace{-0.05in}
    \begin{subfigure}[t]{.18\linewidth}
        \includegraphics[width=\textwidth]{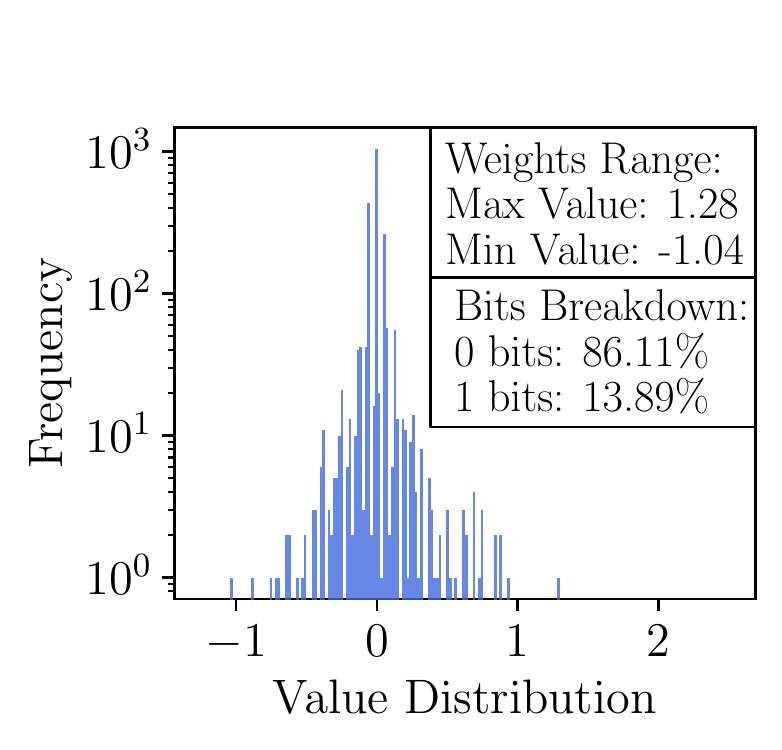}
        \vspace{-0.2in}
        \caption{Weight distribution}
        \label{subfig:NN_weight}
    \end{subfigure}
    \hspace{-0.05in}
    \begin{subfigure}[t]{.16\linewidth}
        \includegraphics[width=\textwidth]{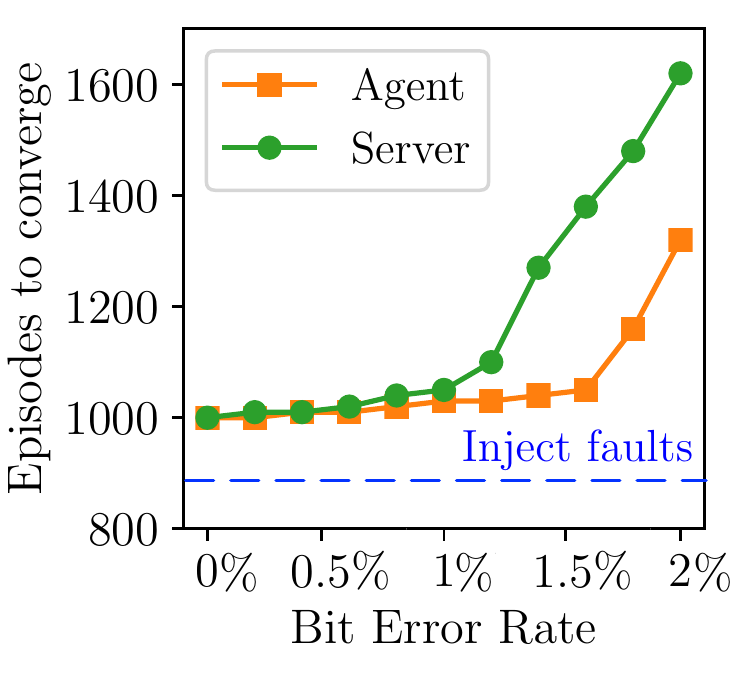}
        \vspace{-0.2in}
        \caption{Policy convergence}
        \label{subfig:NN_transient_convergence}
    \end{subfigure}\\
    \caption{[GridWorld] Transient fault characterization in GridWorld training, including (\blue{a}, \blue{b}) FRL and (\blue{c}) single-agent systems. The fault impact is evaluated by average success rate of all agents within 1000 attempts for each scenario. FRL system is more vulnerable to server faults than agent fault. Single-agent system is more vulnerable to faults than multi-agent system. 0$\rightarrow$1 flip has a higher impact than 1$\rightarrow$0 flip, which can be explained by (\blue{d}) policy weight distribution, where more $0$ bits exist than $1$ bits and policy has a narrow range. (\blue{e}) shows episodes taken to converge after fault injected at the 900th episode. With training more episodes, system can recover from transient faults. }
    \vspace{-0.1in}
    \label{fig:heatmap_multi_gridworld}
\end{figure*}

%

\subsection{FRL Grid-Based Navigation Problem (GridWorld)}
\label{subsec:gridworld}
\subsubsection{Problem Description}
\label{subsubsec:description_gridworld}
\ 
\newline
\indent 
We begin with a simple problem of GridWorld, with $10\times10$ grid mazes (Fig.~\ref{fig:gridworld}). Each cell is characterized into one of four types: \{\textit{hell, goal, source, free}\}. The agent is initialized at \textit{source} and the task is to reach \textit{goal} by avoiding \textit{hell}.
The action space of the agent consists of $\mathcal{A} = \{\textit{up},\textit{down},\textit{right},\textit{left}\}$. At each iteration, the agent observes a one-step state $s\in\mathcal{R}^4$ which corresponds to the nature of the four cells surrounding it. If the corresponding cell is a \textit{hell}, \textit{goal}, or \textit{free}, then state element is -1, 1, or 0 respectively. Hence, we have a finite state space with $|\mathcal{S}|=3^4=81$. 
During each iteration, the agent samples an action from $\mathcal{A}$ and observes a reward based on the next state. The reward is -1, 1, 0.1, or -0.1 if the agent crashed into hell, reached the goal, moved closer to, or moves away from the goal. For each agent $i$, the effectiveness of the policy is quantified by the success rate ($SR_i$) defined by
\begin{equation*}
        SR_i = \frac{\text{\# of times agent $i$ reached goal state}}{\text{total \# of attempts in environment $i$}}  \in [0,1] 
\end{equation*}

The greater the $SR_i$, the better the policy in achieving the goal. The performance of unified policy can be found by taking the average of the success rate across all agents, $SR = \frac{1}{n}\sum_{i=0}^{n-1} SR_i$, where $n=12$ is the number of agents.
Widely used NN-based method is adopted. The policy is quantized to 8-bit without loss of performance given the memory and power constraints of edge devices. We repeat each fault injection campaign 1000 times, which can lead to 95\% confidence level within 1\% error margin.


\subsubsection{Training in FRL GridWorld}
\label{subsubsec:training_gridworld}
\ 
\newline
\indent 
\textbf{Transient faults in training.} Fig.~\ref{subfig:NN_agent}  and Fig.~\ref{subfig:NN_server} demonstrate the impact of faults on GridWorld. It is observed that faults in early episodes with low BER have no effect since the system can recover itself from faults and have enough time to converge (Eq. \ref{eq:convergence}). However, faults occurring after a certain number of episodes with higher BER will degrade the success rate, since faults may destroy the learned model parameters. It is also noted that faults with small BER sometimes result in a slightly higher success rate. We believe this is because hardware bit-flips can be considered as a type of injected noise
that is small enough to maintain a good policy but large enough to regularize model behavior and improve generalization. This is akin to regularization in RL and previous quantization work reports similar impact~\cite{lam2019quantized}.
We also find 0$\rightarrow$1 has a higher impact than 1$\rightarrow$0 flip. This can be explained by Fig.~\ref{subfig:NN_weight}, where 0 bits are more prevalent than 1 bits in the narrow-range policy. Particularly, 0$\rightarrow$1 flip can catastrophically destroy NN policy since outliers in NNs have a much higher impact on the whole system.

\input{tabs/std}

\textbf{Comparing vulnerability of the server and the agents.}
FRL system is more vulnerable to the server's faults than the agents' faults, which can be observed from Fig.~\ref{subfig:NN_agent} and Fig.~\ref{subfig:NN_server}. 
The reason is that faults occurred in an agent will be smoothed by the server; thus, its impact will be reduced. Other agents can help faulty agents quickly recover due to correlations amongst one another. However, faults in server will impact all agents and are equivalent to a randomized policy of all agents to some extend. This motivates us to apply fault mitigation techniques on the server to improve robustness as proposed in \blue{\S}\ref{subsec:fault_mitigation_training}.

\textbf{Single and multi-agent comparison.} Multi-agent FRL system exhibits higher performance and resilience compared to single-agent systems. Fig.~\ref{subfig:single_agent} shows the impact of the fault on a single-agent system (no server). Compared to Fig.~\ref{subfig:NN_server}, it is observed that a multi-agent system can achieve a higher average success rate even under server faults. This is because, in the single-agent system, the policy is only trained for states faced by this agent, which limit its generalization and may lead to overfitting. The agent is prone to failure when the faults make the agent reach unknown states. On the contrary, in the FRL system, the policy is trained on multiple environments by multiple agents and shared with all agents through communication. By optimizing a more collective objective function in Eq. \ref{eq:unified_objective}, the policy can generalize better. 

The generalization capability can be explained by the standard deviation of actions in the policy. A greater standard deviation of the consensus policy indicates a better differentiation between good and bad actions for a given state. Table~\ref{tab:std} shows that in a multi-agent system, the consensus policy has a higher standard deviation, therefore performing better and exhibiting higher resilience than the single-agent system.


\textbf{Policy convergence.}
Transient faults will not affect policy convergence with longer fine-tuning time. We consider convergence as when the unified policy is able to achieve $>$96\% success rate after faults occur, which is an experimental equivalent of Eq.~\ref{eq:convergence}.
We inject bit-flips towards the end of training (900th episode) with different BERs and measure when the policy can fully recover. Fig.~\ref{subfig:NN_transient_convergence} indicates that the system can finally recover from faults as long as trained longer. 

\begin{figure}[t!]
\centering
    \includegraphics[width=.75\columnwidth]{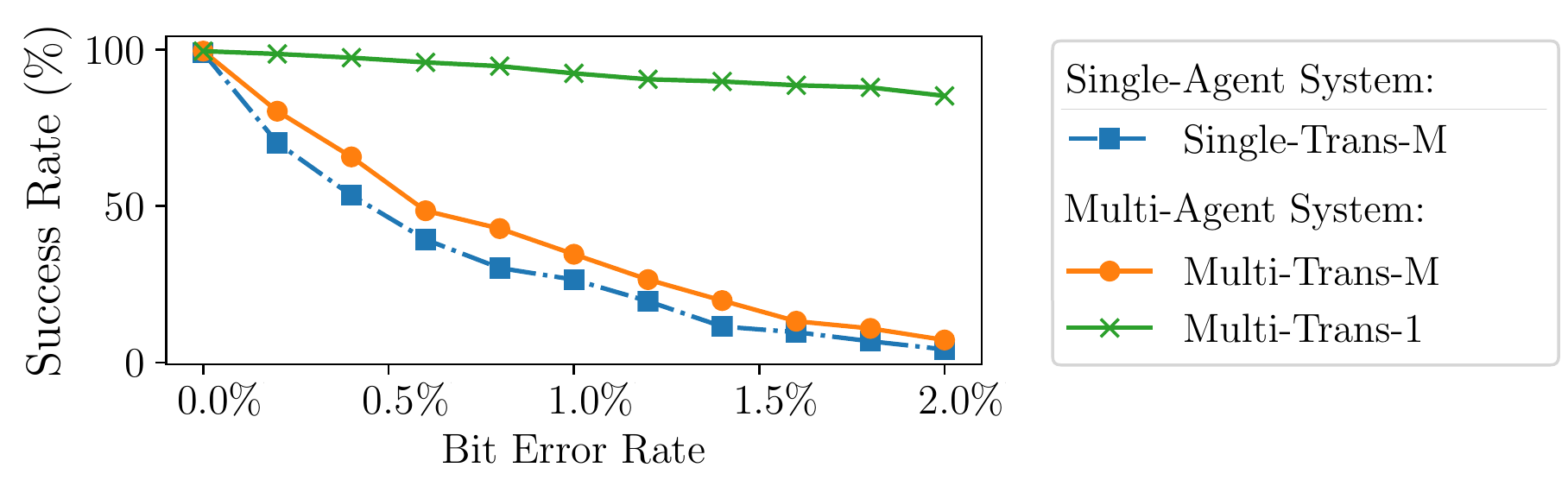}
    \caption{[GridWorld] Transient fault characterization in GridWorld inference. Fault can impact only one action step (Multi-Trans-1) or whole sequential decision-making steps (Multi-Trans-M). Multi-agent system is more resilient to transient fault than single-agent system.}
    \vspace{-0.2in}
    \label{fig:inference}
\end{figure}

\subsubsection{Inference in FRL GridWorld}
\label{subsubsec:inference_gridworld}
\ 
\newline
\indent 
\textbf{Transient fault in inference.}
During inference, each agent utilizes the unified policy and consistently conducts exploitation. This is an iterative procedure where the policy is used in each action based on the current state. Transient faults can happen in the read register (Multi-Trans-1) that only affects one action step, or in memory (Muti-Trans-M) which affects all the following actions. Fig.~\ref{fig:inference} demonstrates that Multi-Trans-1 has a negligible impact on the performance. This can be explained by the sequential decision-making procedure of FRL where faults in one action step may be corrected by following actions and does not necessarily result in task failure.


\textbf{Single and multi-agent comparison.}
Multi-agent system is more resilient to transient fault than single-agent system. Comparing Multi-Trans-M and Single-Trans-M in Fig.~\ref{fig:inference}, we observe the average success rate of all agents in a multi-agent system is consistently higher than the single-agent system. The reason is that policy trained in the multi-agent system can generalize better and react correctly to more states. 
\subsection{FRL Drone Navigation Problem (DroneNav)}
\label{subsec:drone}
\subsubsection{Problem Description}
\label{subsubsec:description_drone}
\ 
\newline
\indent 
We further experiment on a more complex problem of drone autonomous navigation. Drone is a complex
system~\cite{krishnan2021machine}, and we use PEDRA~\cite{anwar2020autonomous} as the drone navigation platform which is powered by Unreal Engine and AirSim. PEDRA contains various 3D realistic environments and the objective of the drone is to navigate across the environments after being initialized at a starting point. 
There is no goal position, and the drone is required to avoid the obstacles as long as it can. At each iteration $t$, the drone captures an RGB image from front-facing camera which is taken as the state $s_t \in \mathbb{R}^{(320\times 180 \times 3)}$. An action $a_t\in \mathcal{A}$ is taken based on $s_t$. We consider a perception based probabilistic action space with 25 actions ($|\mathcal{A}|=25$). A depth-based reward function is designed to encourage the drone to stay away from obstacles. A NN-based policy with three Conv layers and two FC layers is used to estimate the action probabilities based on states.

The policy is first trained offline using REINFORCE algorithm and then fine-tuned online using transfer learning. We consider four drones in FRL system. Faults are injected in a single random step with various BER during online fine-tuning or inference. Experiments are repeated 100 times for each case. We use safe flight distance to quantify the performance which is the average distance traveled by the drone before collision. 




\subsubsection{Training in FRL Drone Autonomous Navigation}
\label{subsubsec:training_drone}
\begin{figure}[t!]
\centering
    \begin{subfigure}[t]{.34\linewidth}
        \includegraphics[width=\textwidth]{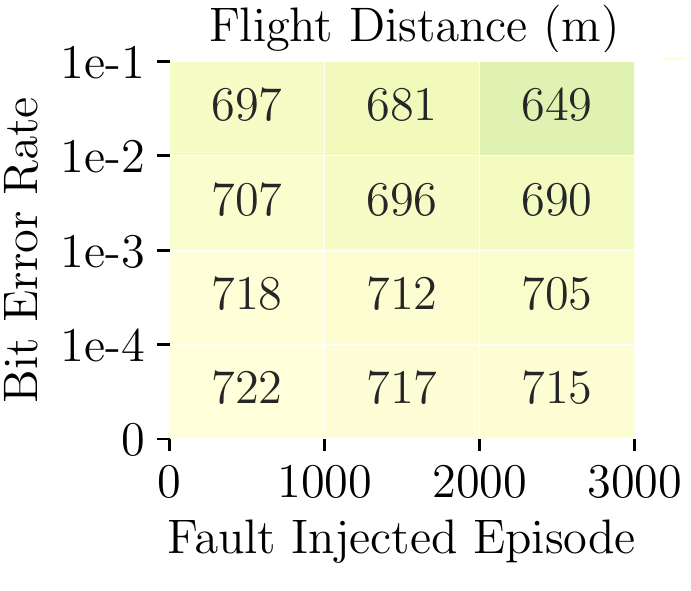}
        \vspace{-0.2in}
        \caption{Agent fault}
        \label{subfig:FedRL_drone_training_agent}
    \end{subfigure}
    \begin{subfigure}[t]{.27\linewidth}
        \includegraphics[width=1\textwidth]{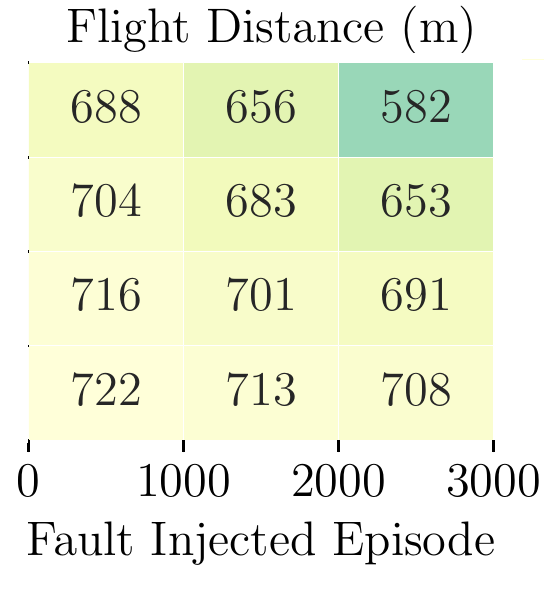}
        \vspace{-0.2in}
        \caption{Server fault}
        \label{subfig:FedRL_drone_training_server}
    \end{subfigure}
    \begin{subfigure}[t]{.35\linewidth}
        \includegraphics[width=\textwidth]{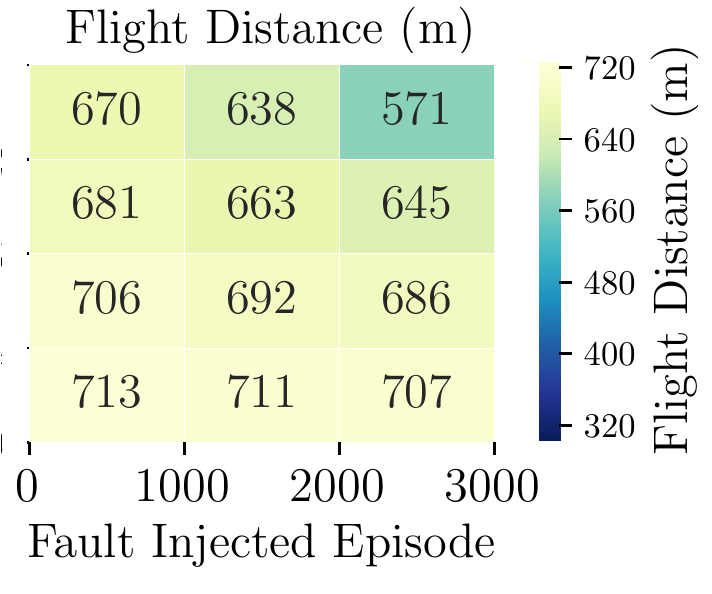}
        \vspace{-0.2in}
        \caption{Single-drone system}
        \label{subfig:single_drone_training}
    \end{subfigure}
    \caption{[DroneNav] Fault characterization in drone navigation training, including (\blue{a}, \blue{b}) FRL and (\blue{c}) single-drone systems. The fault impact is evaluated by average safe flight distance (the longer, the better). The system is more vulnerable to server faults. FRL system is more robust than single-drone system.
    }
    \vspace{-0.1in}
    \label{fig:drone_FI}
\end{figure}

\ 
\newline
\indent 
\textbf{Fault types and fault locations.}
Fig.~\ref{subfig:FedRL_drone_training_agent} and Fig.~\ref{subfig:FedRL_drone_training_server} show the impact of transient faults on FRL drone navigation system. Faults that occurred in later fine-tuning episodes with a higher BER impact the system more. For system components, server fault has a higher impact than agent fault. These trends are similar to our observations in GridWorld.

\textbf{Single and multi-drone comparison.}
Multi-drone systems exhibit higher resilience and performance than a single-drone system. Comparing Fig.~\ref{subfig:single_drone_training} with Fig.~\ref{subfig:FedRL_drone_training_agent} and Fig.~\ref{subfig:FedRL_drone_training_server}, we observe that drones in FRL system achieve a higher average safe flight distance. This is because correctly operating drones can help faulty drones quickly recover through communication and sharing a global policy. NN policy trained in a multi-drone system can generalize better and deal with more unknown states, making it also more resilient to server faults.

\begin{figure}[t!]
\centering
        \begin{subfigure}[t]{.45\linewidth}
        \includegraphics[width=\textwidth]{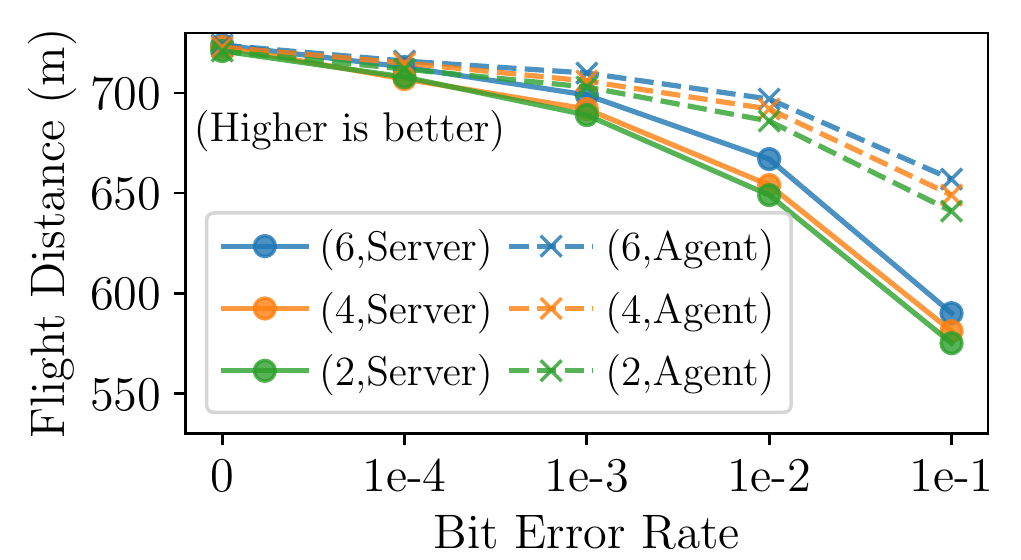}
        \vspace{-0.2in}
        \caption{Different number of drones}
          \label{subfig:drone_num_training}
    \end{subfigure}
    \hspace{-0.08in}
    \begin{subfigure}[t]{.54\linewidth}
        \includegraphics[width=\textwidth]{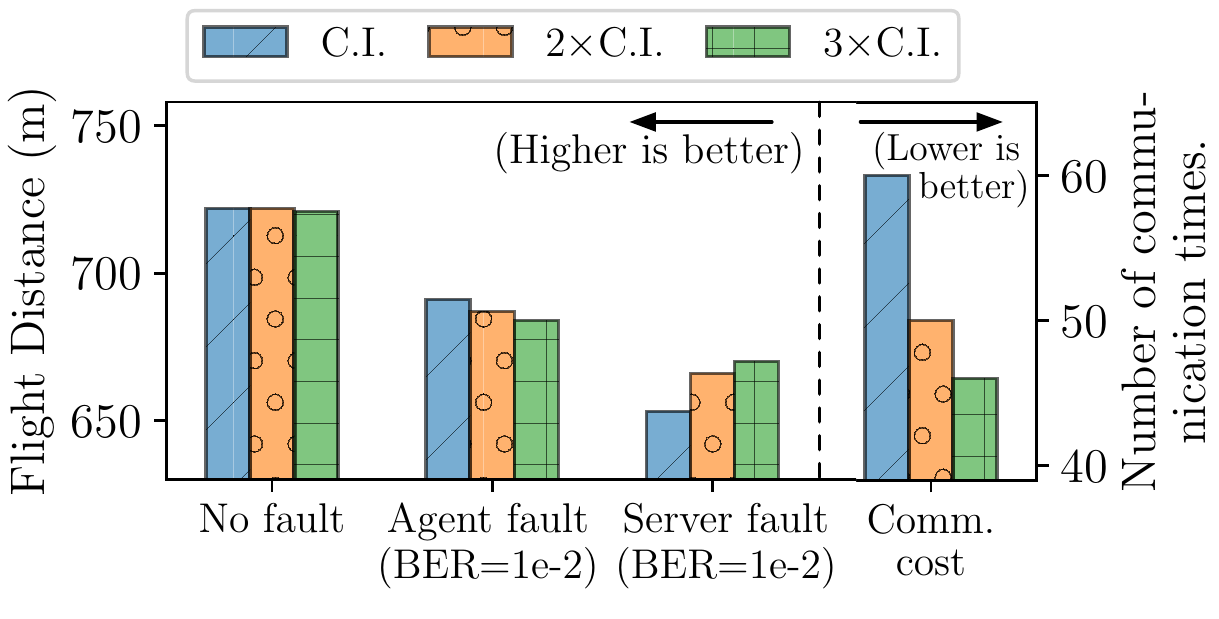}
        \vspace{-0.2in}
        \caption{Different communication intervals}
          \label{fig:comm_interval}
    \end{subfigure}
    \caption{[DroneNav] \blue{(a)} System resilience under various drone numbers. (6/4/2, server/agent) means (total number of drones, faults location). More drones can improve system reliability. \blue{(b)} System resilience under various communication interval (\textit{C.I.}). \textit{C.I.} is doubled by 2$\times$ and 3$\times$ after 2000th episode. Longer interval can reduce communication cost and server fault impact, while increase agent fault impact.}
    \vspace{-0.2in}
    \label{fig:num_drone}
\end{figure}

\textbf{Number of drones.}
More drones helps improve resilience (Fig.~\ref{subfig:drone_num_training}). When faults occur in agents, more drones bring a stronger smoothing effect via server that alleviates fault impact. More normal drones can help faulty drone recover quickly by sharing their knowledge. When faults occur in server, the swarm of drones makes the policy generalize better with more acquired information, helping drones find alternative, feasible actions.

\textbf{Communication Interval.}
Various communication intervals between the server and the agents exhibit a trade-off between resilience and communication cost. In the current setup, after the 2000th episode, we increase the communication interval by 2$\times$ and 3$\times$ since drones usually perform more exploitation. Interestingly, as shown in Fig.~\ref{fig:comm_interval}, during this period, a higher communication interval makes the system more vulnerable to the agent's faults. This is because a higher interval means that the number of times that the faulty drone received correct information is lowered. However, higher communication interval alleviates the impact of server faults, since a fewer number of communications indicate less computation in the server and therefore a lower probability of the server to transmit faulty data to the agents. By increasing the interval by 3$\times$ after the 2000th episode, total communication cost reduces by 23.3\%.


\subsubsection{Inference in FRL Drone Autonomous Navigation}
\label{subsubsec:inference_drone}
\ 
\newline
\indent 
\textbf{Data types.}
The reliability depends on the data type. We test three fixed-point data types Q(sign, integer, fraction): Q(1,4,11), Q(1,7,8) and Q(1,10,5) as a case study. We find that Q(1,10,5) is the most vulnerable because it provides an unnecessarily large range for value representation that results in large deviations when faults do occur. By contrast, Q(1,4,11) is more robust and fits the model better, indicating that a data type which can optimally capture the parameter range can improve resilience. This is in line with some recent works~\cite{reagen2018ares,tambe2020algorithm,wan2021analyzing}.
\subsection{Summary and Takeaways}
\label{subsec:summary_FedRL}
Transient faults impact FRL navigation system to varying degrees. Faults occurred in later training episodes have higher impacts. Server faults have higher impacts than agent faults. Multi-agent system exhibits higher resilience and higher policy performance than single-agent system. More agents make the system more robust. We reveal the trade-off between cost and resilience under various server-agent communication intervals. Different layers and data types exhibit various resilience, depending on layer topology, position, and representation range. 

%% file: tabs/std.tex
\begin{table}[t!]
\centering
\caption{[GridWorld] Standard deviation (\textit{std}) of the consensus policy. Larger std indicates better differentiation between good and bad actions. Multi-agent system has higher \textit{std} than single-agent system, indicating its higher performance and resilience. $n$ means \#agents. 
}
\vspace{-2pt}
\resizebox{\columnwidth}{!}{%
\huge
\begin{tabular}{c|cccc}
\hline
                  &  \textbf{Single-agent} & \textbf{Multi-agent (n=4)} & \textbf{Multi-agent (n=8)} & \textbf{Multi-agent (n=12)} \\ \hline
\textbf{\textit{Std}} &  0.255 & 0.405 & 0.472 &  0.504             \\
\hline   
\end{tabular}
\label{tab:std}
}
\vspace{-0.18in}
\end{table}

%% file: 5mitigation.tex
\section{Fault Detection and Recovery Techniques}
\label{sec:fault_mitigatioin}
Taking FRL system characteristics into consideration, we present cost-effective fault mitigation techniques for both training (\blue{\S}\ref{subsec:fault_mitigation_training}) and inference (\blue{\S}\ref{subsec:fault_mitigation_inference}) with overhead evaluation.
\subsection{Training: Server Checkpointing}
\label{subsec:fault_mitigation_training}
\textbf{Fault detection.}
Based on observations that faults that occur during training will result in a drop in the reward, we use the agent's cumulative reward drop as an indicator of the fault. If the reward drop in any agent exceeds $p\%$ for $k$ consecutive episodes ($p$ and $k$ are parameters), we assume faults are detected in the system. If only one agent (i.e., $ag_i$) has a reward drop, we assume that faults have occurred in the agent. If more than half of the agents' rewards drop, we conclude that the fault has occurred in the server. Note that we use an application-level metric, instead of a conventional bit-level comparison for fault detection. This is motivated by our observation that faults with low BER do not necessarily degrade final performance due to the long-term nature of the decision making process and the inherent collaborative feature of the FRL system.

\textbf{Fault recovery.}
We observe the faults in the server have more severe impact than faults in the agents (Fig.~\ref{fig:drone_FI}). So we propose to save a checkpoint for the model weight in the server and update it at every 5 communication intervals. Once faults are detected in one agent $ag_i$, the checkpoint will be copied from server to $ag_i$. If the reward of $ag_i$ goes back to being normal after $k$ episodes, we deduce that the fault is transient and the system has recovered. Similarly, once faults are detected in the server, we revert the server's weights to the last checkpoint and continue the training. If agent's rewards increase to normal after $k$ episodes, we infer that faults have been suppressed. 

\begin{figure}[t!]
\centering
    \begin{subfigure}[t]{.45\linewidth}
        \includegraphics[width=\textwidth]{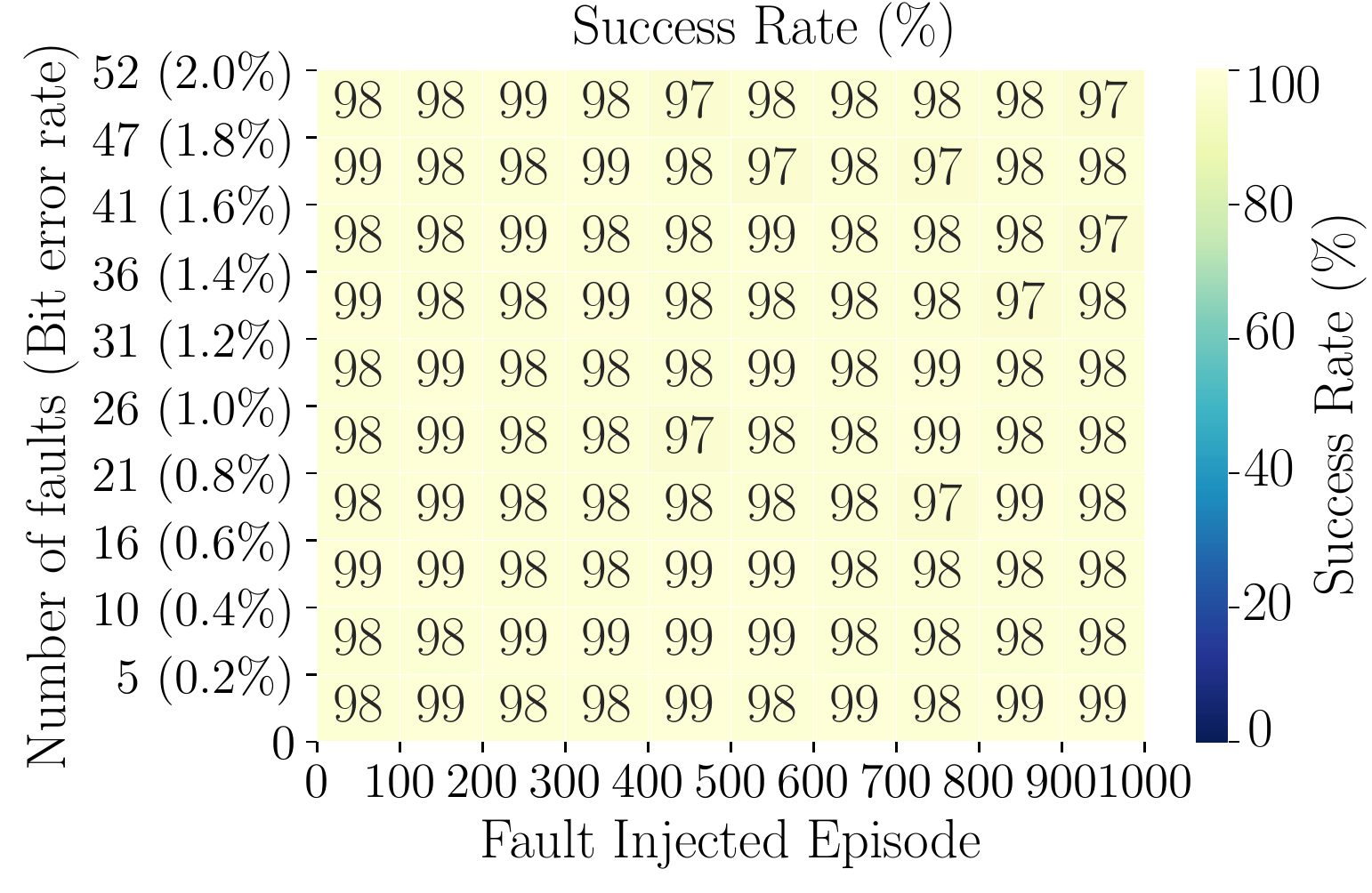}
        \vspace{-0.2in}
        \caption{FRL GridWorld}
        \label{subfig:mitigation_training_gw}
    \end{subfigure}
    \hspace{0.1in}
    \begin{subfigure}[t]{.4\linewidth}
        \includegraphics[width=\textwidth]{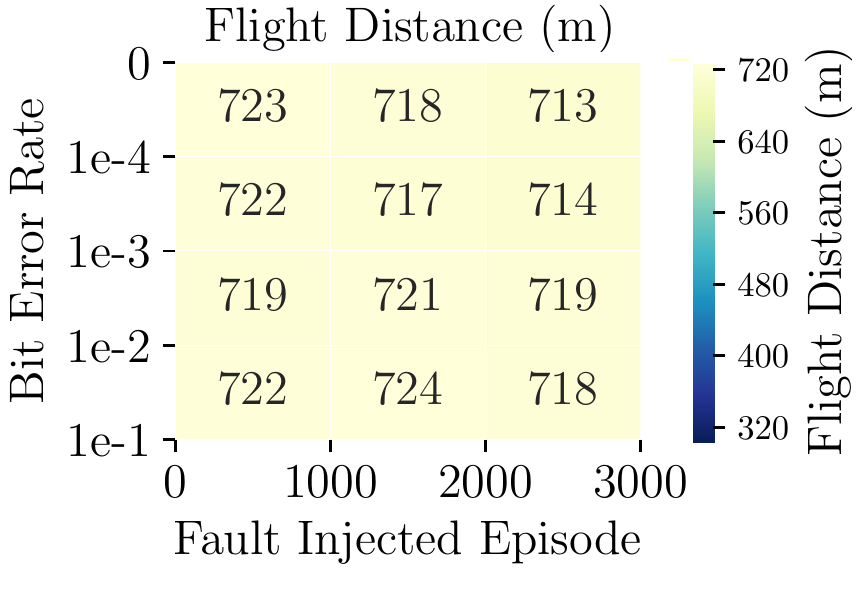}
        \vspace{-0.2in}
        \caption{FRL drone navigation}
        \label{subfig:mitigation_training_drone}
    \end{subfigure}
    \caption{[Fault D\&R] The effect of fault detection and recovery in training using server checkpointing. Compared with Fig.~\ref{fig:heatmap_multi_gridworld} and Fig.~\ref{fig:drone_FI}, agents success rate and drones flight distance recover to near baseline.}
    \vspace{-0.1in}
    \label{fig:mitigation_training}
\end{figure}

\textbf{Evaluation.}
We evaluate the fault recovery on GridWorld ($p=25$, $k=50$) and drone navigation ($p=25$, $k=200$). The parameters are chosen adaptable to test scenarios. Fig.~\ref{fig:mitigation_training} shows that the success rate of agents maintains $>$96\% in GridWorld and flight distance reaches $>$712$m$ in drone navigation after applying the fault mitigation scheme. Note that server checkpointing is performed asynchronously with its self-updating and communication, bringing no runtime overhead. When the server is stopped for memory diagnosis, other components are not impacted and will continue to execute.

\subsection{Inference: Range-Based Anomaly Detection}
\label{subsec:fault_mitigation_inference}
\textbf{Fault detection and recovery.} Based on the observation that values with high magnitude are more likely to be outliers and have higher impact, we leverage range-based  anomaly  detection, which is similar as~\cite{li2017understanding,wan2021analyzing}. Before the agents start to conduct steady exploitation, the weights of each layer will be tallied and its range will be derived as ($w_{min}$, $w_{max}$), and a 10\% margin is applied on detector as ($1.1w_{min}$, $1.1w_{max}$). If any weight is outside this range, the false alarm will be triggered. Once the fault is detected, the operations around this value will be skipped. The rationale is that a small value has higher probability to become an outlier if bit-flips occurred. NN has inherent sparsity and most values are around zero.

\begin{figure}[t!]
\centering
    \begin{subfigure}[t]{.45\linewidth}
        \includegraphics[width=\textwidth]{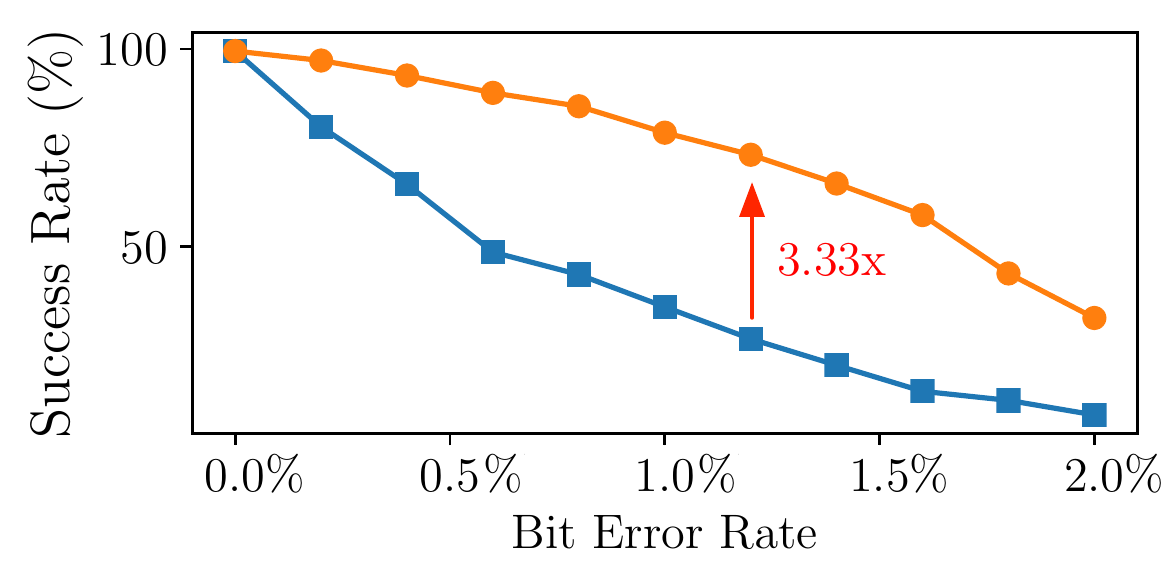}
        \vspace{-0.2in}
        \caption{FRL GridWorld}
        \label{subfig:mitigation_inference_gw}
    \end{subfigure}
    \begin{subfigure}[t]{.45\linewidth}
        \includegraphics[width=\textwidth]{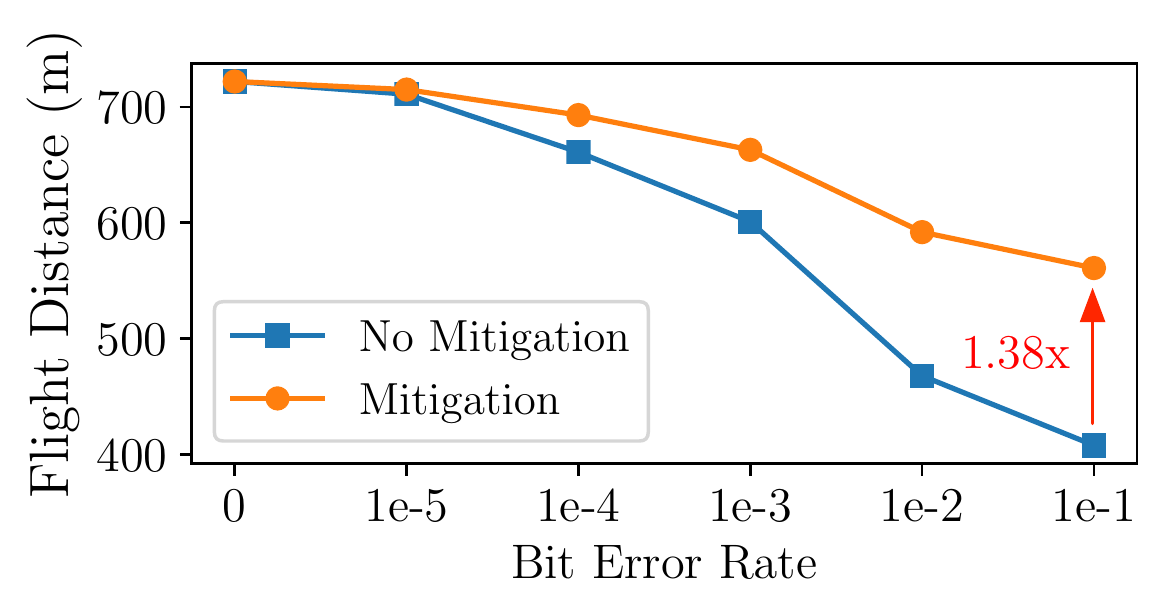}
        \vspace{-0.2in}
        \caption{FRL drone navigation}
        \label{subfig:mitigation_inference_drone}
    \end{subfigure}
    \caption{[Fault D\&R] The effect of fault detection and recovery in inference using range-based anomaly detection. Agent success rate and drone flight distance are improved 3.3$\times$ and 1.4$\times$ under faults.}
    \vspace{-0.1in}
    \label{fig:mitigation_inference}
\end{figure}

\textbf{Evaluation.}
We evaluate the effectiveness of anomaly detection in both GridWorld and drone navigation inference (Fig.~\ref{fig:mitigation_inference}). Faults are injected in NN weights. Compared with no-fault mitigation scenarios, the average success rate of agents and average safe flight distance of drones are improved 3.3$\times$ and 1.4$\times$, respectively, after applying the mitigation scheme.


\newcommand{\txtab}{
\renewcommand{\arraystretch}{1.2}
\resizebox{0.5\linewidth}{!}{
\begin{tabular}{|c|c|c|}
\hline
                     & \textbf{AirSim Drone} & \textbf{DJI Spark} \\ \hline
Type         & mini-UAV                & micro-UAV                   \\ \hline
Size (mm) & 650               & 170                   \\ \hline
Weight (g)         & 1652               & 300                 \\ \hline
\begin{tabular}[c]{@{}c@{}} Battery \\ Capacity (mAh) \end{tabular}      & 6250               & 1480                 \\ \hline
\end{tabular}
}
}
\begin{figure}[t!]
    \subfloat{\includegraphics[valign=B, width=0.48\linewidth]{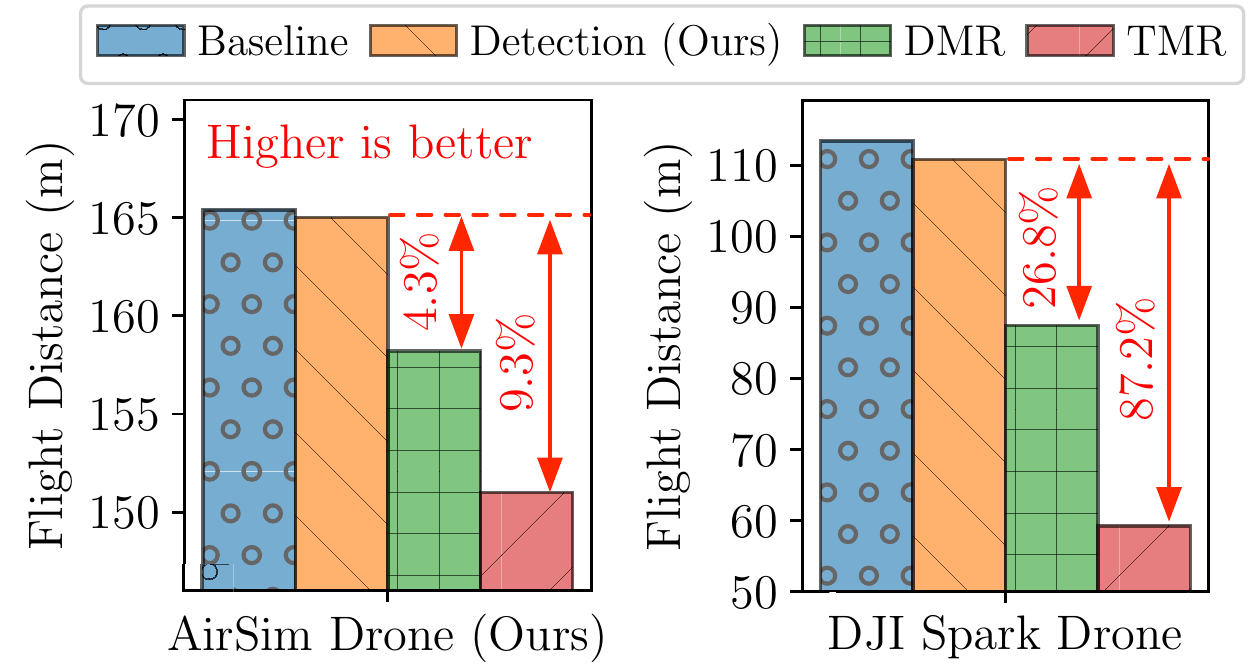}}
    \subfloat{\adjustbox{width=0.48\columnwidth,valign=B,raise=\baselineskip}\txtab}
    \caption{[Fault D\&R] Comparison of redundancy-based approaches (DMR and TMR) and the proposed fault detection and recovery scheme from end-to-end drone systems' performance perspective.}
    \label{fig:overhead_compare}
    \vspace{-0.2in}
\end{figure}



\textbf{Compute overhead.} We adopt a drone performance analysis model~\cite{krishnan2020sky,wan2021roofline} to evaluate the end-to-end overhead of our proposed scheme and compare with redundancy-based hardware protections (DMR and TMR). Two types of drones, AirSim drone and the DJI Spark (with the same settings as~\cite{krishnan2020sky}), are used as platforms. Our scheme incurs $<$2.7\% runtime overhead with negligible drone performance degradation (Fig.~\ref{fig:overhead_compare}). However, TMR degrades the safe flight distance by 9.3\% on AirSim drone and 87.8\% on DJI Spark compared to our scheme. The rationale is that hardware redundancy brings higher power and weight, thus lowering flight velocity and safe flight distance. This further corroborate that lightweight application-aware protection techniques are needed for resource-constrained systems.

%% file: 6conclusion.tex
\vspace{-0.03in}
\section{Conclusion}
\vspace{-0.03in}
\label{sec:conclusion}
Practical reliability considerations for swarm intelligence require a better understanding of reliability in end-to-end distributed navigation systems.
We present FRL-FI, a fault analysis to
characterize the impact of transient faults on both training and inference stages via in-depth fault injection experiments. 
We present cost-effective fault detection and recovery schemes by checkpointing at the server and detecting anomalous values on the agents, which improves system reliability by up to 3.3$\times$.

\vspace{-0.03in}

\section*{Acknowledgements}
This work was supported in part by C-BRIC and ADA, two of six centers in JUMP, a Semiconductor Research Corporation (SRC) program sponsored by DARPA.
\vspace{-0.03in}